\documentclass[letterpaper, 10 pt, conference]{ieeeconf}  % Comment this line out if you need a4paper

\IEEEoverridecommandlockouts 

\usepackage{graphicx}
\usepackage{mathtools}
\usepackage{comment}
\usepackage[bottom]{footmisc}
\usepackage{cite}
\usepackage{stfloats}
\usepackage{multirow}
\usepackage{subfigure}
\usepackage[ruled,vlined]{algorithm2e}
\title{\LARGE \bf
MobileCharger: an Autonomous Mobile Robot with Inverted Delta Actuator for Robust and Safe Robot Charging
}
\author{\authorblockN{Iaroslav Okunevich, Daria Trinitatova, Pavel Kopanev, and Dzmitry Tsetserukou}
\authorblockA{ \textit{Skolkovo Institute of Science and Technology, Moscow, Russia 121205}\\
\{iaroslav.okunevich, daria.trinitatova, pavel.kopanev, d.tsetserukou\}@skoltech.ru}}

\begin{document}

\IEEEoverridecommandlockouts

\pubid{\makebox[\columnwidth]{978-1-7281-2989-1/21/\$31.00 \copyright2021 IEEE\hfill} \hspace{\columnsep}\makebox[\columnwidth]{ }}

\maketitle
%\thispagestyle{empty}
%\pagestyle{empty}

%%%%%%%%%%%%%%%%%%%%%%%%%%%%%%%%%%%%%%%%%%%%%%%%%%%%%%%%%%%%%%%%%%%%%%%%%%%%%%%%
\begin{abstract}
MobileCharger is a novel mobile charging robot with an Inverted Delta actuator for safe and robust energy transfer between two mobile robots. The RGB-D camera-based computer vision system allows to detect the electrodes on the target mobile robot using a convolutional neural network (CNN). The embedded high-fidelity tactile sensors are applied to estimate the misalignment between the electrodes on the charger mechanism and the electrodes on the main robot using CNN based on pressure data on the contact surfaces. Thus, the developed vision-tactile perception system allows precise positioning of the end effector of the actuator and ensures a reliable connection between the electrodes of the two robots. The experimental results showed high average precision (84.2\%) for electrode detection using CNN. The percentage of successful trials of the CNN-based electrode search algorithm reached 83\% and the average execution time accounted for 60 $s$. MobileCharger could introduce a new level of charging systems and increase the prevalence of autonomous mobile robots.

\textit{Index terms} -- Autonomous mobile robots, computer vision, intelligent sensors and actuators, deep learning methods, factory automation.
\end{abstract}

%%%%%%%%%%%%%%%%%%%%%%%%%%%%%%%%%%%%%%%%%%%%%%%%%%%%%%%%%%%%%%%%%%%%%%%%%%%%%%%%
\section{Introduction}

Currently, the field of mobile robots application is incredibly wide and includes retail \cite{petrovsky2020customer}, inspection \cite{peel2018localisation}, logistics \cite{gao2018multi,fragapane2020autonomous}, agriculture \cite{karpyshev2021autonomous}, disinfection \cite{ultrabot}, etc. The limitations of mobile robots in terms of operation time \cite{batterySchedule2020} are related to their energy capacity. To achieve full autonomy in indoor environment, e.g. in factory, the robots have to cover large areas of the operation time, which leads to the requirement of an increased operation time of the robot. This factor can be compensated by a larger number of installed docking stations. However, for outdoor robots that explore an unknown area, the problem of limited operation time is crucial.

\begin{figure}[t]
\centering
\includegraphics[width=1\linewidth]{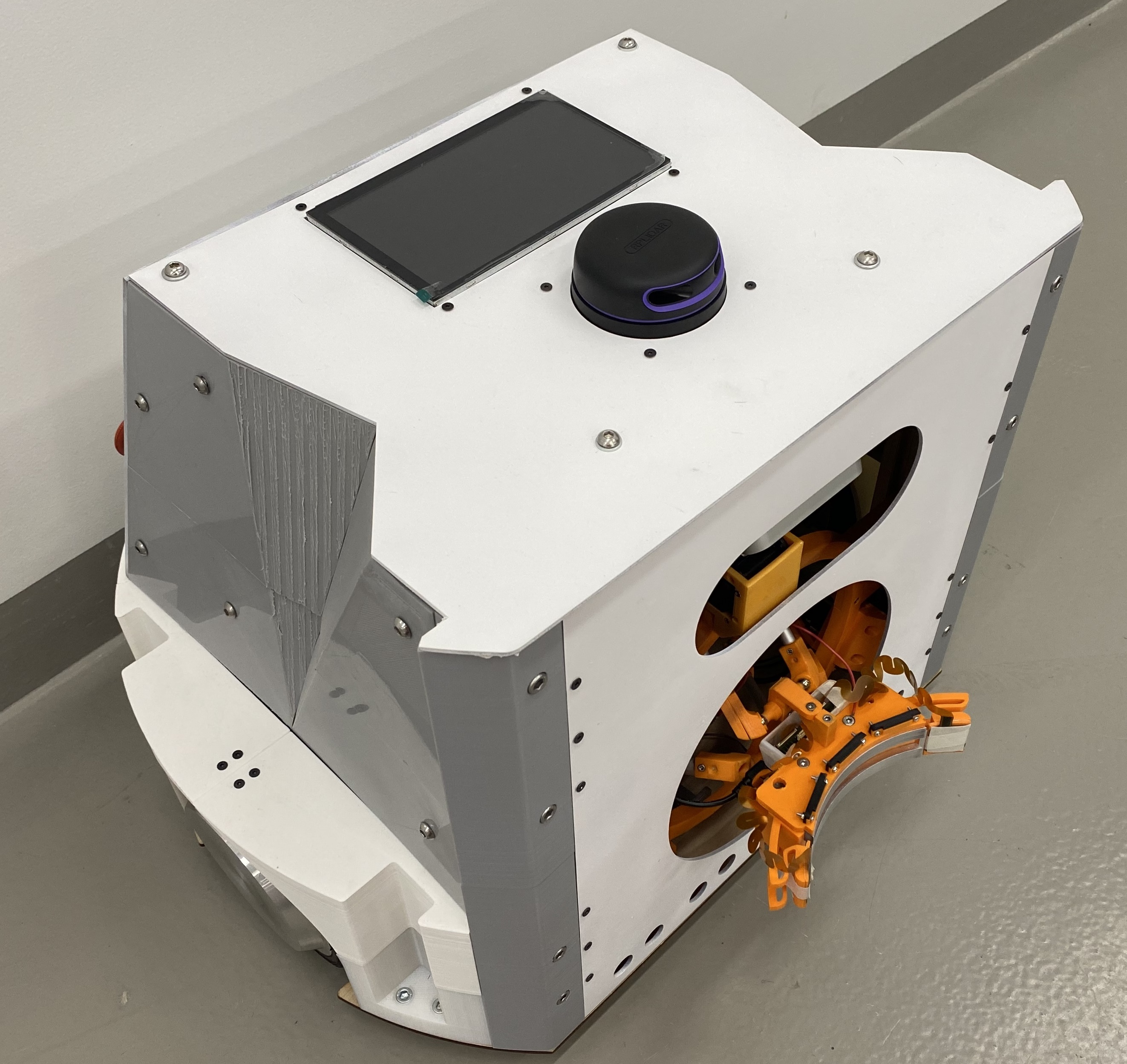}
\caption{The prototype of MobileCharger.}
\label{MC}
\vspace{-1em}
\end{figure}

Existing strategies to increase the operation time include boosting the battery capacity of the robot or applying renewable energy sources. Although increasing the capacity of the battery seems to be one of the most straightforward solutions, it significantly affects the weight and size of the mobile robot. Another approach is based on the use of renewable energy sources, such as solar panels \cite{plonski2016environment}, which could lead to an unlimited working range of the robot. However, such systems tend to be sensitive to external conditions and have a low charging rate. An alternative approach could be taken from the aviation industry. The aircraft can be refueled from the tanker to increase its operation time. Similarly, an additional mobile robot, which would be cheaper and simpler than the original one, could recharge the main robot to achieve long-term deployments. 

\subsection{Problem statement}
\pubidadjcol %“pulls up” the text in the second column to prevent it from blindly running into the publication ID.
The autonomous mobile charging robot has to detect the electrodes on the target robot, move to the electrode area, and charge the main robot. The charging process has a restriction on the relative position between robots. The electrodes on the charger and main robot should be parallel or tilted at a slight angle to each other for successful charging. To accomplish this task, the following requirements were formed:

\begin{itemize}
     \item \textit{Detection of electrodes and estimation of their coordinates in the actuator coordinate system}: The charging robot has to detect the electrodes on the main robot and to estimate their coordinates for successful docking.
    \item \textit{Workspace of the actuator}: The actuator needs to have at least 3 degrees of freedom (DoF) with a move range of more than 60 $mm$ in each direction. It allows to compensate for the positioning error between two robots during charging. 
    \item \textit{Measurement of the misalignment between electrodes of two robots}: The main requirement for charging is a reliable and safe connection between electrodes. The angular misalignment between the electrodes on the charger and the target robot leads to a short circuit. The horizontal or vertical misalignment leads to a decrease in the contact area of the electrodes, which can cause damage to them. The charging robot has to have a sensing system to solve the misalignment issue.
    \item \textit{Transferring energy}: The target robot is charged during the 4 hours by the voltage of 22 $V$ from the charger. The actuator has to transmit energy continuously for at least 4 hours with the same voltage.
\end{itemize}

\subsection{Related works}

Currently, there are several prototypes of autonomous auto-charging robots \cite{AuCS, CDAC,CDLC}. The most advanced technology is using a mobile autonomous system for charging. For example, there are prototypes of mobile robots for recharging electric vehicles in public places \cite{AMCPP}. The most famous examples are fully autonomous robots ``Laderoboter"\footnote{https://www.volkswagen-newsroom.com/en/stories/volkswagen-lets-its-charging-robots-loose-5700} by Volkswagen with a robotic arm and ``EVAR"\footnote{https://www.evar.co.kr/eng/pro04.php} by Samsung spin-off.

To perform charging, a safe and reliable connection between the electrodes of two robots has to be reached. This can be achieved by using various types of actuators. Applying a robotic arm as an actuator helps to reach the main robot at a distance of about 0.2--1.2 $m$. However, this solution significantly affects the weight, cost, and power supply of the charging robot. Another approach is the use of a Cartesian manipulator, e.g., as in the ``EVAR" robot. The main downside of this solution is the large space requirement. An Inverted Delta actuator invented by us is a more appealing option since it is compact, easy to manufacture, and at the same time has a high operational speed. 

The sensor system of the charging robot has to detect the relative position of the electrodes to prevent a short circuit during charging. There are several types of sensors that could be applied to solve this problem. For example, the array of retroreflective photoelectric sensors was utilized to align the electric vehicle with the wireless charging station \cite{photoelectric}. The disadvantage of this technology is the need to install additional receivers on the main robot. Another approach is to use a camera-based vision system. The relative position of the electrodes can be calculated from the color and depth images of RGB-D camera. This method is applied in the harvester robot for fruit pose estimation \cite{harvester}. The use of such a system in the charging robot will allow detecting the main robot and its electrodes. However, the close proximity during the charging prevents the visual system from measuring the relative position between the electrodes. An alternative approach is to develop a tactile sensing system for the robot. Tactile sensors are applied for different tasks in robotics \cite{tactileSensors}. They do not require the installation of an additional system on the main robot and can estimate the relative position of the electrodes using machine learning methods.

In the current work, we propose MobileCharger, an autonomous mobile robot with the Inverted Delta actuator with semicircular electrodes for autonomous charging of the mobile robots (Fig. \ref{MC}). For precise positioning of the charger's end effector and providing a robust connection between the electrodes of two robots, we developed a CNN-driven vision-tactile perception system.

\section{Design of MobileCharger}

\subsection{System configuration}

\begin{figure}[t]
\centering
\includegraphics[width=0.95\linewidth]{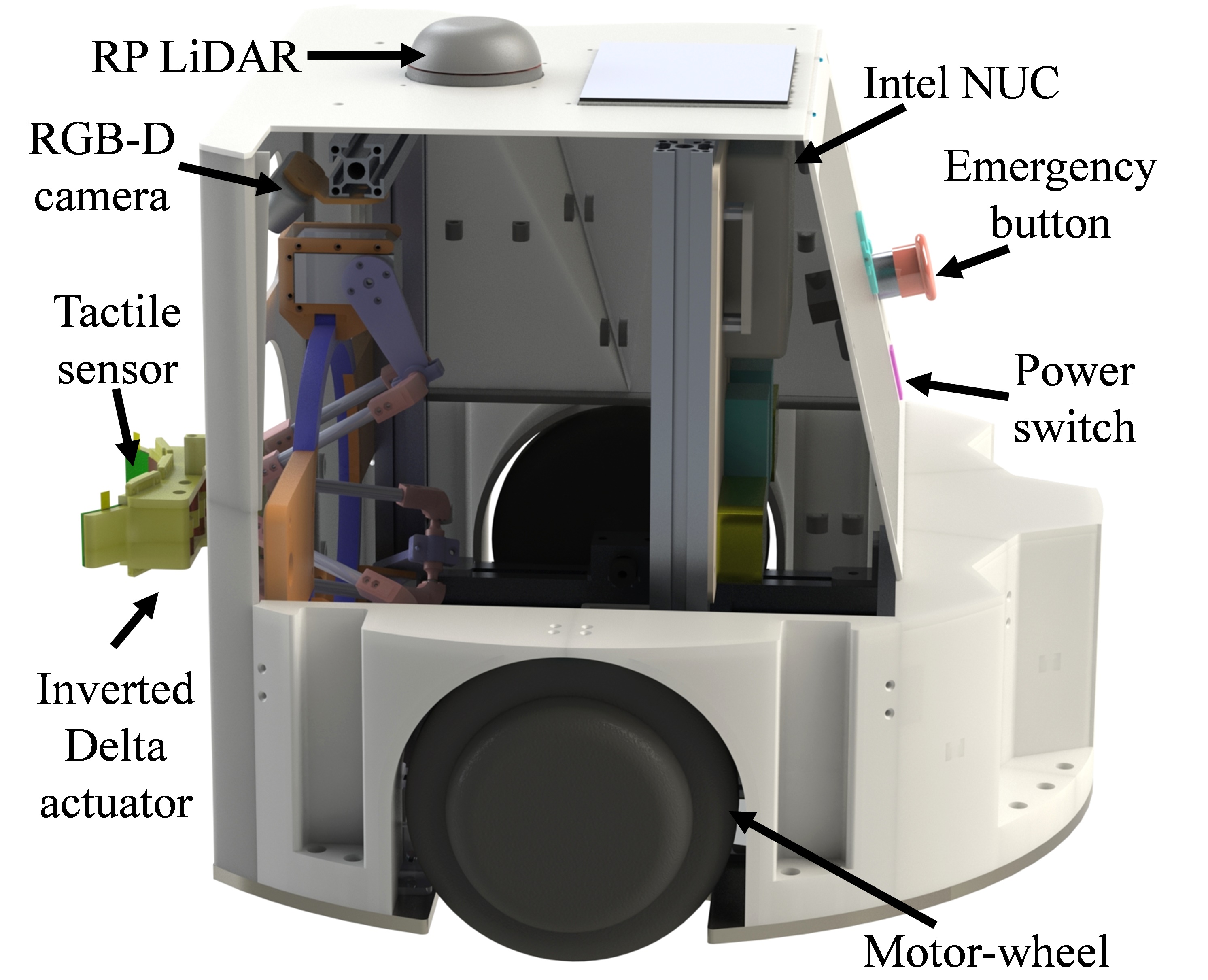}
\caption{Decomposition of the robot structure.}
\label{DMC}
\vspace{-0.5em}
\end{figure}

The elements of MobileCharger are shown in Fig. \ref{DMC}. The architecture of MobileCharger system consists of 3 blocks. The Computing Unit block comprises high and low-level controllers, RGB-D camera, and LiDAR. The high-level controller is Intel NUC7i5BNK embedded computer. OpenCM 9.04 Dynamixel controller is applied as a low-level controller because it has specific ports for connecting to the Dynamixel servomotors, which are used to actuate the Inverted Delta mechanism. The communication between the Intel NUC and the microcontroller was implemented using the UART protocol. The RGB-D camera and LiDAR are connected to the computer via a USB interface. The computer solves the inverse kinematics problem for the Inverted Delta mechanism, analyses data from sensors, and sends the desired angular velocities of wheels to the motor controllers. The velocity value is transmitted via the RS-485 interface. 

The second block is the Inverted Delta mechanism proposed by us. It is used as the actuator and has a setup ring, servomotors, and end effector with electrodes. OpenCM board controls servomotors through UART. The end effector is equipped with tactile sensors supplied by 5 VDC power source. The tactile sensor data is transmitted to the controller via an analog interface.

\begin{figure}[t]
\centering
\includegraphics[width=0.9\linewidth]{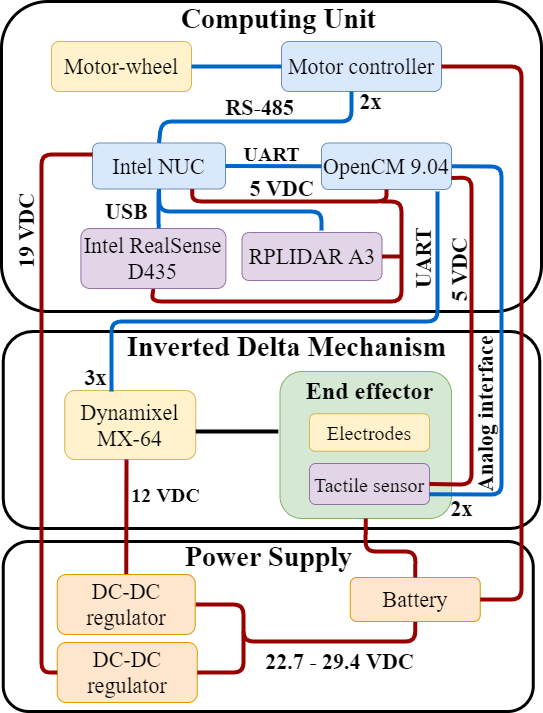}
\caption{The system configuration of MobileCharger.}
\label{UBFD}
\vspace{-1em}
\end{figure}

Power Supply block consists of a battery and two DC-DC regulators. The output voltage of the battery equals 22.7 - 29.8 VDC. Servomotors and Intel NUC require 12 VDC and  19 VDC, respectively. Thus, DC-DC regulators are applied to achieve these voltages. Motors and electrodes in the end effector are electrified by the battery. The system configuration of MobileCharger is shown in Fig. \ref{UBFD}.

\subsection{Embedded sensors}
MobileCharger has 3 groups of sensors: LiDAR, \mbox{RGB-D} camera, and tactile pressure sensors. The Intel RealSense D435\footnote{https://www.intelrealsense.com/depth-camera-d435/} RGB-D camera is applied to detect electrodes on the main robot and to estimate their coordinates. RPLIDAR A3\footnote{https://www.slamtec.com/en/Lidar/A3} is used as a primary sensor for robot navigation and localization. High-fidelity tactile sensor arrays are integrated into the end effector of the Inverted Delta actuator to provide the misalignment perception between electrodes of the charger and main robot. 

\subsection{Actuator structure}
\begin{figure}[t]
     \centering
  \includegraphics[width=0.85\linewidth]{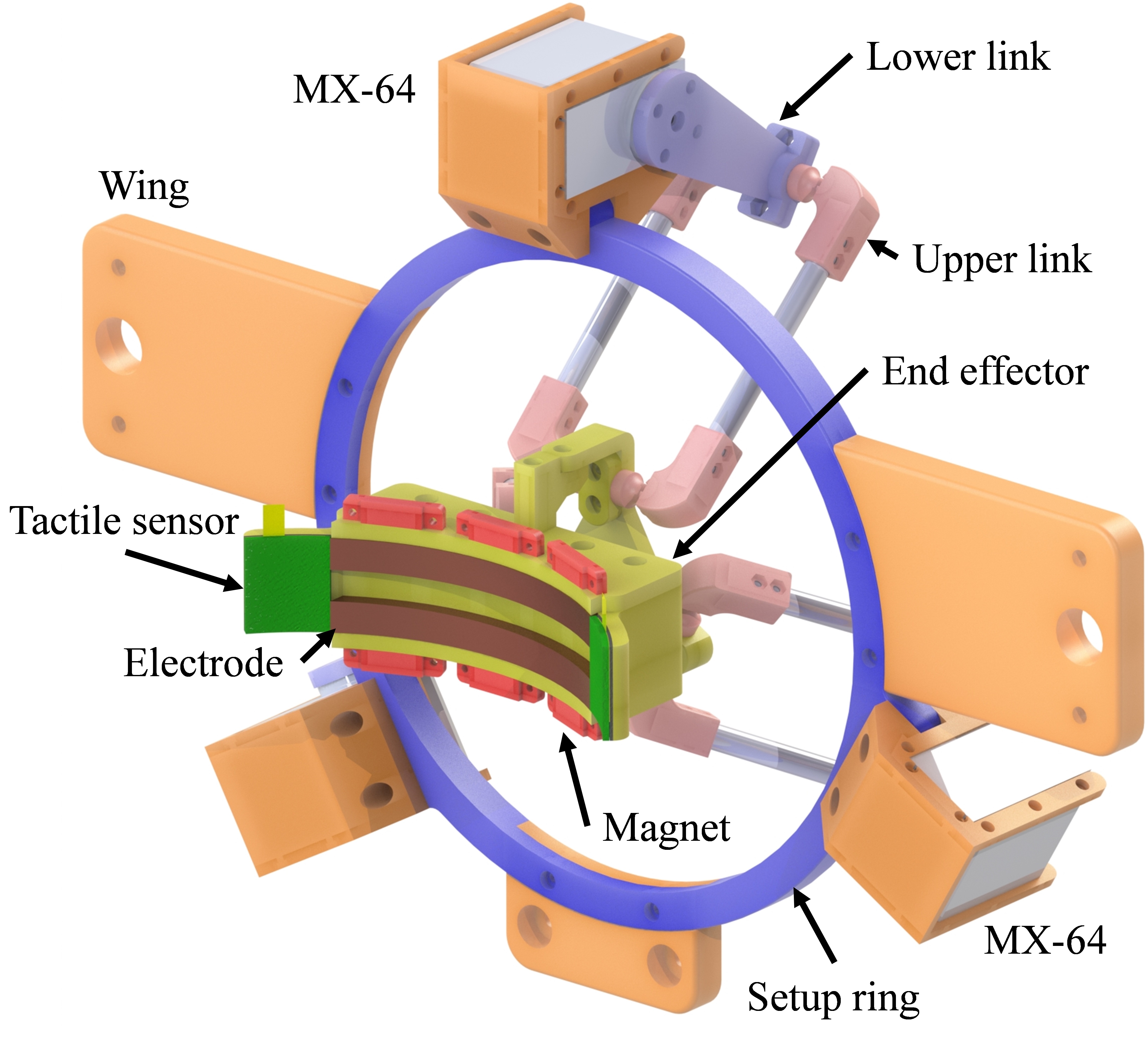}
   \caption{A CAD model of DeltaCharger.}
   \label{DeltaCad}
 \vspace{-1em}
 \end{figure}

We have developed DeltaCharger \cite{DC}, the Inverted Delta robot, as an actuator to achieve high precision with a small size (Fig. \ref{DeltaCad}). In contrast to traditional Delta robot structure, the proximal and distal links occupy same working area achieving a compact design. The Inverted Delta mechanism is dubbed vDelta (in\textbf{v}erted or \textbf{v}ersus Delta) \cite{deltaTouch}.  DeltaCharger comprises a setup ring with three servomotors (Dynamixel MX-64) and a moving platform with two electrodes. The setup ring is attached to the mobile charging robot with three wings. The movement of the end effector is achieved by means of three interchangeable kinematic limbs of the actuator. 

The end effector consists of a platform with two duralumin electrodes, six magnets, and two tactile sensor arrays. The electrodes can withstand a current of 18 $A$. The magnets keep the electrodes of the charger in constant contact with the discharged main robot. Tactile sensors are applied to detect a misalignment between the electrodes of the end effector and electrodes of main robot. The tactile perception system based on these sensors is described in section \ref{perception}. With the weight of 274 $g$, the end effector can generate a strong force of 25 $N$ in the normal direction. The specification of DeltaCharger is presented in the Table \ref{tabl:T1}.

\begin{table}[ht]
    \centering
    \caption{Specification of DeltaCharger}
    \label{tabl:T1}
    \begin{tabular}{|c|c|}
    \hline
     Servomotors & \shortstack{\\Dynamixel MX-64}  \\
    \hline
     Weight of DeltaCharger & \shortstack{\\938 $g$} \\
     \hline
     Weight of the end effector & \shortstack{\\274 $g$} \\
    \hline
     Material & \shortstack{\\PLA, aluminium} \\
    \hline
    Working range along the $X$, $Y$ axes & \shortstack{\\120 $mm$} \\
    \hline
    Working range along the $Z$-axis & \shortstack{\\110 $mm$} \\
    \hline
    \end{tabular}
\end{table}

\section{Operational Principle}
This section presents the strategy of the robot's behavior and describes the operation of the visual and tactile perception systems.

\subsection{Strategy of operation}
The behavior strategy assumes that the robotic system consists of the main robot and autonomous charging robot. The main discharged robot sends the information about its current coordinates to MobileCharger. When the charge level of the main robot drops below the threshold value, the main robot will stop the current operation and send the request for recharging to the charging robot. In indoor environment, the transmission of information would be applied using \mbox{Wi-Fi} communication. In outdoor environment, the applying of radio waves is the most relevant\cite{RadioRobots}. MobileCharger potentially can be applied in both conditions.

When MobileCharger accepts the request, it will move to the area of obtained coordinates and start to search for the electrodes using the RGB-D camera. On the one hand, the charging robot does not have the map, which the main robot has. However, MobileCharger knows its coordinates relative to the initial position of the main robot. It can build the path to the obtained coordinates. LiDAR helps with navigation and collision avoidance. However, these tasks are not covered in this paper. 

The charging robot explores the circular area with a radius of 0.5 $m$ centered at the point of the obtained coordinates. If it can not detect the electrodes, the radius of the circle will be increased to 2.5 $m$ with 0.5 $m$ increment. If the charger still cannot find the electrodes, it should return to the initial position. In the case of electrode detection, the charger moves to the electrodes and RGB-D camera estimates their position. The vDelta reaches this position and docks with electrodes of the target robot. After that, the tactile perception system has to estimate the misalignment between the electrodes of the robots using tactile sensors. If the misalignment angle is lower than the critical one, the charging robot starts to charge the main robot. Otherwise, it should return to the initial position.

\subsection{System for electrode position detection}

MobileCharger has to detect the electrodes on the target mobile robot. The RGB data from the RGB-D camera is applied for object detection. Pose estimation of electrodes is calculated using point cloud, which is calculated by depth data from the RGB-D camera. 

We applied a neural network with convolutional layers as an object detection algorithm. We chose an open-source pre-trained CNN YOLOv3 based on Darknet neural network with 106 layers and retrained it on our dataset \cite{redmon2018yolov3}. 

\begin{figure}[!t]
\begin{center}
\subfigure[\label{original_images}Source images.]{
\includegraphics[width=0.98\linewidth]{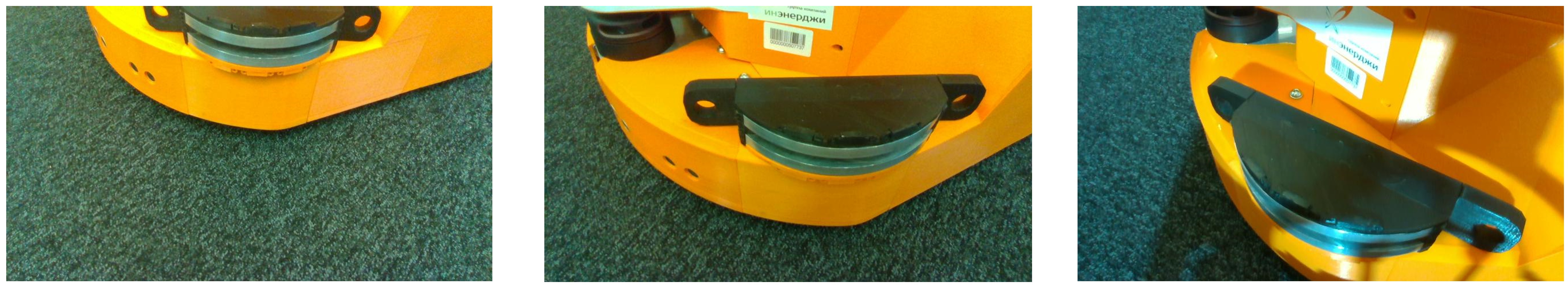}}
\subfigure[\label{CV_images}Detected electrodes.]{
\includegraphics[width=0.98\linewidth]{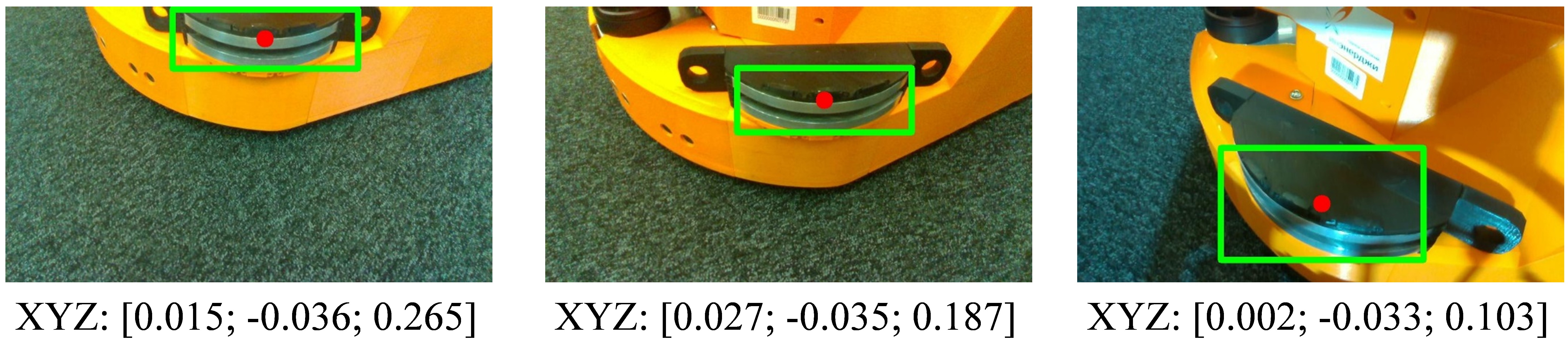}}
\caption{Results of CNN-based electrode detection on the test dataset. Red dots represent the coordinates of the centers of bounding boxes given in the coordinate system of DeltaCharger.}\label{ExampleCV}
\end{center}
\vspace{-1em}
\end{figure}

We collected a dataset where the electrodes were located on different parts of the main robot and the position of MobileCharger was changed relative to the main robot. To improve the robustness of the detection system, we varied the illumination of the electrodes in the dataset. Our dataset consists of 170 samples, 120 in train and 50 in test sets. The images were made in three different lighting conditions: bright, normal, and dark. In addition, the distance between the robots varied from 5 to 40 $cm$. The yaw angle of the MobileCharger varied from 75$^{\circ}$ to 105$^{\circ}$. Each sample in the dataset has RGB and depth images, a Boolean variable, which shows if the image contains electrodes or not, and the distance between the charging robot and target electrodes. The resolution of the images was 840x480 pixels. The image dataset was annotated manually using the LabelImg\footnote{https://github.com/tzutalin/labelImg} graphical image annotation tool. The source images with detection results of the applied CNN are presented in Fig. \ref{ExampleCV}.

\begin{figure}[h]
    \centering
    \includegraphics[width=0.85\linewidth]{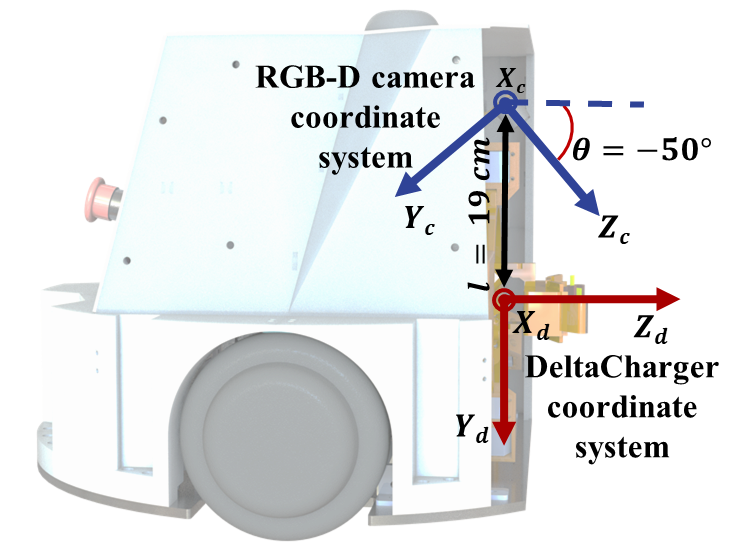}
    \caption{Position of the RGB-D camera coordinate system relative to the DeltaCharger coordinate system.}\label{CS}
\vspace{-0.5em}
\end{figure}

The detection system returns the bounding box with the electrode. The position of the center of the bounding box is calculated from the point cloud. In some cases this point has no depth value. The reason for this is the limitations of the depth camera. If the depth of a point equals zero, the system starts to search the nearest pixel with non-zero depth. The obtained center of the bounding box in the camera coordinate system is used to estimate the position of the electrodes. To calculate actuated angles of vDelta for the required position, the position of electrodes is recalculated in the DeltaCharger coordinate system. The center of the camera is higher than the center of DeltaCharger. The direction of the $X$-axis of both coordinate systems is the same. In addition, the camera coordinate system is tilted along the pitch axis ($X$-axis) by 50$^{\circ}$ relative to the DeltaCharger coordinate system (Fig. \ref{CS}).

The coordinates of the electrodes in the DeltaCharger coordinate system are calculated using the following transformation matrix:

\begin{equation}\label{eq:TransfromMatrix}
\begin{pmatrix}
        X_d\\
        Y_d\\
        Z_d\\
        1\\
    \end{pmatrix}=
    \begin{pmatrix}
        1&0&0&0\\
        0&\cos \theta&-\sin \theta&l\\
        0&\sin \theta&\cos \theta&0\\
        0&0&0&1\\
    \end{pmatrix}
    \begin{pmatrix}
        X_c\\
        Y_c\\
        Z_c\\
        1\\
    \end{pmatrix},
\end{equation}
where $X_c, Y_c, Z_c$ are the coordinates of electrodes in the camera coordinate system, $\theta$ is the angle of rotation around the $X$-axis (pitch angle), which equals $-50^{\circ}$, $l$ is the translation along the $Y$-axis, equals -19 $cm$, and $X_d, Y_d, Z_d$ are the coordinates of electrodes in the DeltaCharger coordinate system. Upon successful electrode detection, the charger moves to the electrodes according to the Algorithm \ref{algorithm}. 

\begin{algorithm} [h] 
\caption{\emph{Electrode search algorithm}}\label{algorithm}
    \While {the electrodes are not reachable}{
        detect the electrodes on the RGB image\;
        
        \eIf{the electrodes are detected}{
        
        calculate the position of the centre of bounding box\;
        
        \If{the bounding box is not at the centre of the image}{
                rotate around the yaw axis by 1$^{\circ}$\;
            }
            \eIf{the bounding box is closer than 11 $cm$}{
                move forward 4 $cm$\;
                
                the electrodes are reachable\;
                
                end while loop\;
            }{
                move forward 1 $cm$\;
             }
            
        }{
        \eIf{it is not a 5th detection attempt}{
            
                try to detect the electrodes one more time
            }{
            the electrodes are not in the field of the camera view\;
            
            end while loop\;
            }
            }
    }
\end{algorithm}

First, MobileCharger rotates around the yaw axis in 1$^{\circ}$ increments until the electrodes are centered on the RGB-D camera image. After that, it moves forward with 1 $cm$ increments and repeats the rotation step. When the distance to the center of the bounding box is less than 11 $cm$, MobileCharger performs the last rotation cycle, moves forward 4 $cm$, and completes the search algorithm. 
A distance of 11 $cm$ is the maximum distance along the $Z$-axis that the actuator can reach to dock with the electrodes on the main robot. The final movement of 4 $cm$ allows to compensate for the error of detecting the position of electrodes along the $Z$-axis.
After final movement, we subtract 4 $cm$ from the $Z$ coordinate of the center of the bounding box in the DeltaCharger coordinate system. If MobileCharger can approach the electrodes within 11 $cm$ and they remain visible, the algorithm is completed successfully. The electrodes can be lost from the field of view if MobileCharger rotates at a very large angle or CNN cannot detect the electrodes from the image. In this case, MobileCharger tries to detect electrodes on the current image in 5 attempts because the electrodes detection may have different results from the same position. Failure to detect the electrodes after 5 trials means that MobileCharger has no electrodes in the field of view, and the trial is considered to be failed. 

\subsection{Tactile perception system}\label{perception}

Mobile robots often perform their tasks on uneven surfaces that can create some issues for the charging process due to the mismatch of the electrodes between the two robots. The electrodes of the charger end-effector should be parallel to the ones on the main robot, and each electrode of the charger should be in contact with only one electrode of the main robot. If the main robot is tilted, one electrode can contact two other electrodes, causing a short circuit. To avoid this problem, the misalignment angle between electrodes can be measured. If it is lower than a threshold value, charging can take place. In addition, the error of the electrode detection by the vision system in proximity to the target robot equals 1 $cm$. To overcome this limitation, we can measure the horizontal and vertical misalignments of the end effector after the docking to correct its position.

\begin{figure}[h!]
    \centering
    \includegraphics[width=0.90\linewidth]{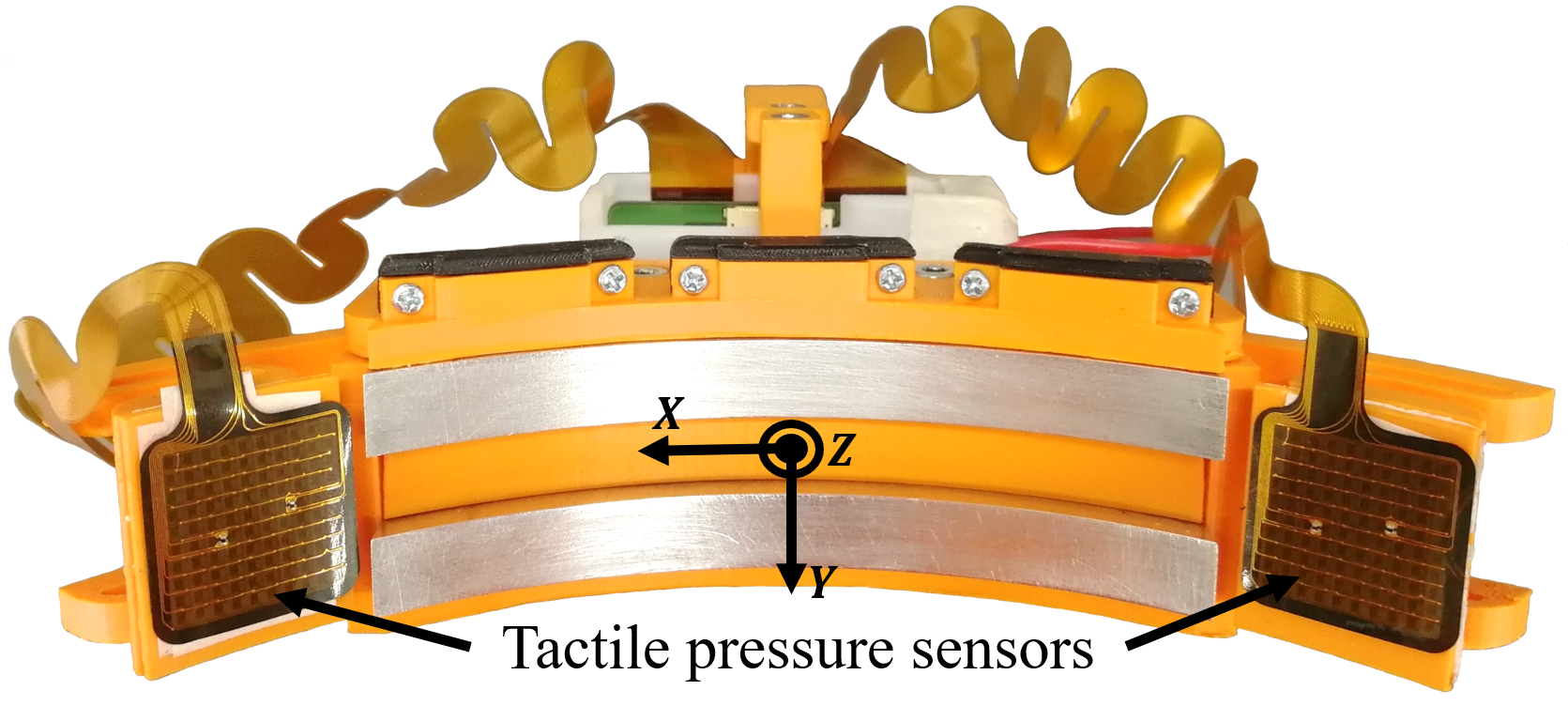}
    \caption{End effector with embedded tactile sensor arrays.}
    \label{Sensor}
\end{figure}

In our system, we propose to use tactile sensors since they are one of the best instruments to collect data in a close proximity environment \cite{PHDtactileSensor}. The end effector of vDelta was embedded with tactile sensor arrays \cite{yem2019}. Each array is capable of sensing the maximum frame area of $5.8\ cm^2$ with a resolution of 100 points per frame (Fig. \ref{Sensor}). The sensing frequency is 120 $Hz$ (frames per second). The sensors allow the system to precisely detect the pressure on the edges of the charger terminal. The force detection range of sensors is from 1 to 9 $N$. 

\begin{figure}[t!]
    \centering
    \includegraphics[width=0.98\linewidth]{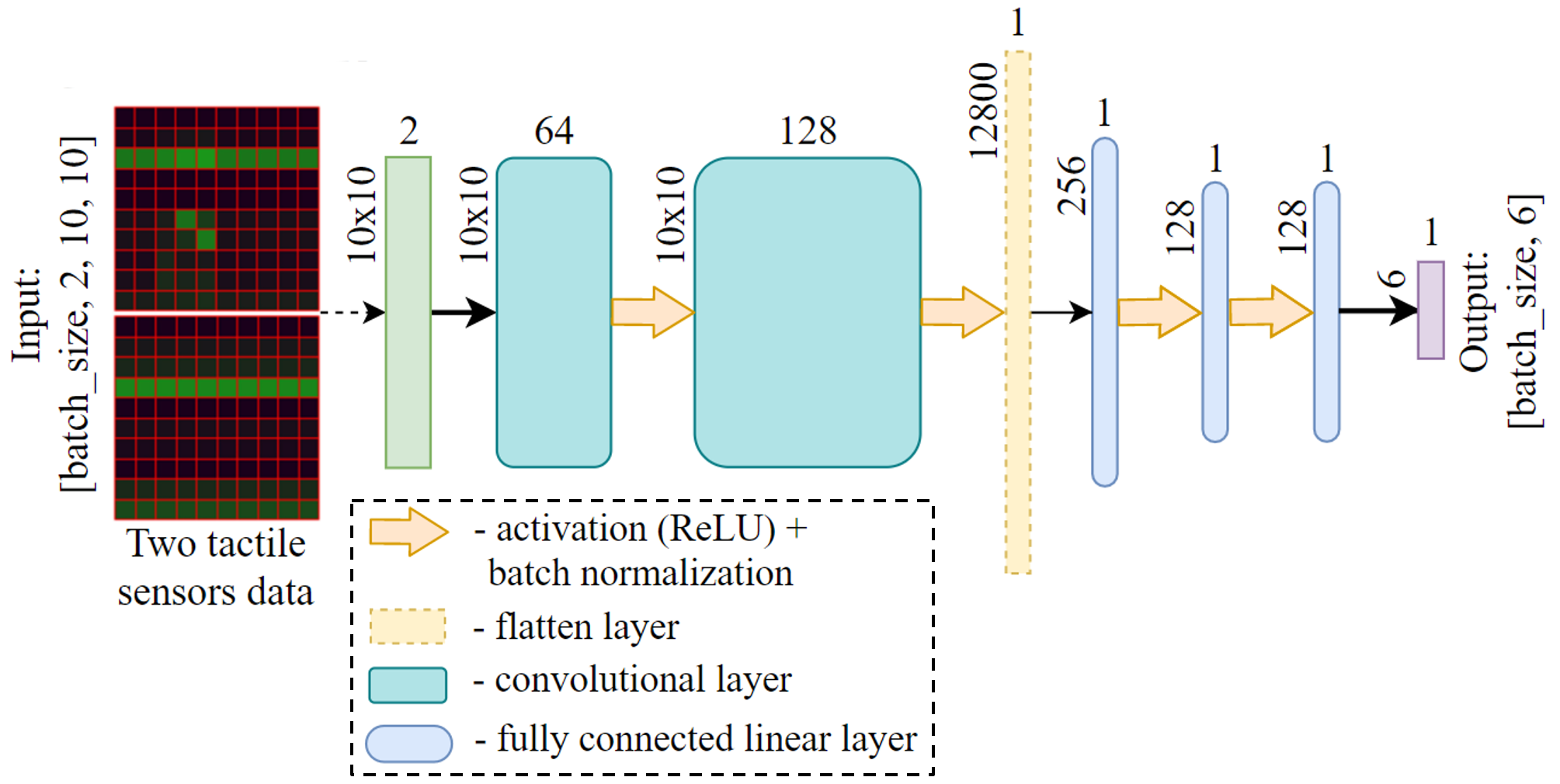}
    \caption{CNN architecture for tactile perception system.}
    \label{CNN}
\vspace{-0.5em}
\end{figure}

To estimate the electrode misalignment, we formulated a classification problem  \cite{DC}. To solve it, we have developed a CNN with 2 convolutional layers with the kernel of 3x3 and 3 fully connected linear layers with ReLU nonlinear activation functions and batch normalization (Fig. \ref{CNN}). We collected two datasets for training and validation of implemented CNN. 

\begin{figure}[!h]
 \begin{center}
 \subfigure[Tactile patterns for different misalignment angles $\varphi$.]{
 \includegraphics[width=0.95\linewidth]{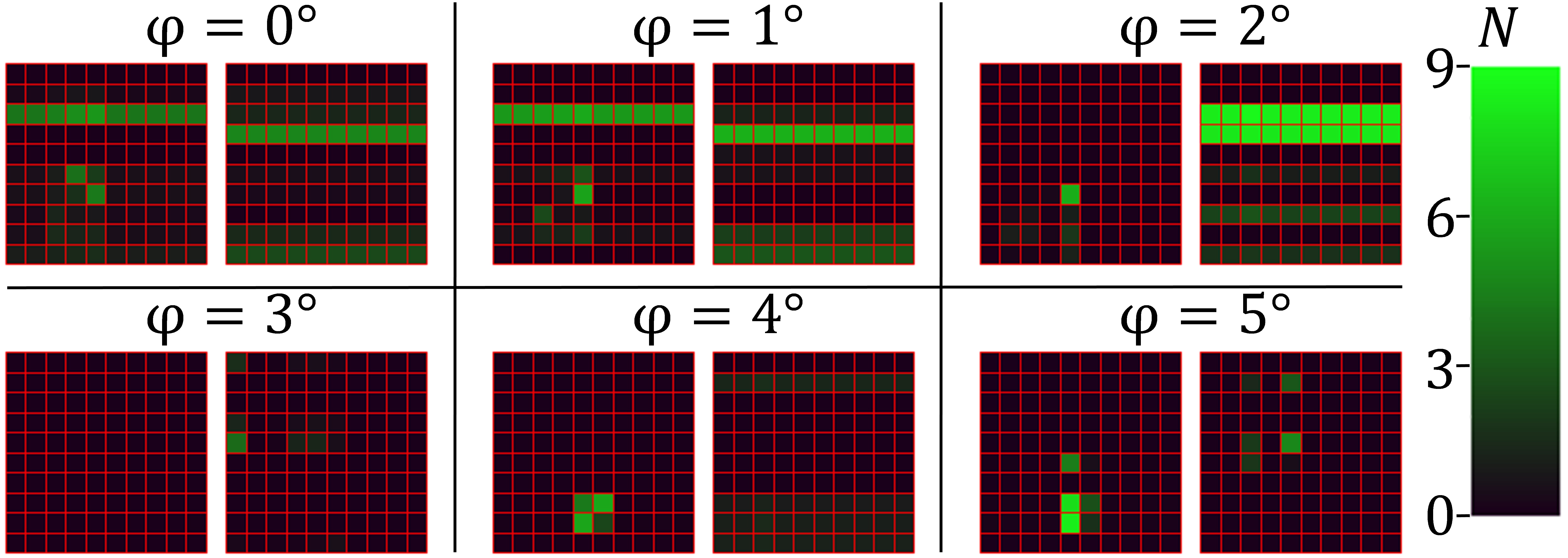}}
 \subfigure[Tactile patterns for different horizontal and vertical misalignments.]{
 \includegraphics[width=0.99\linewidth]{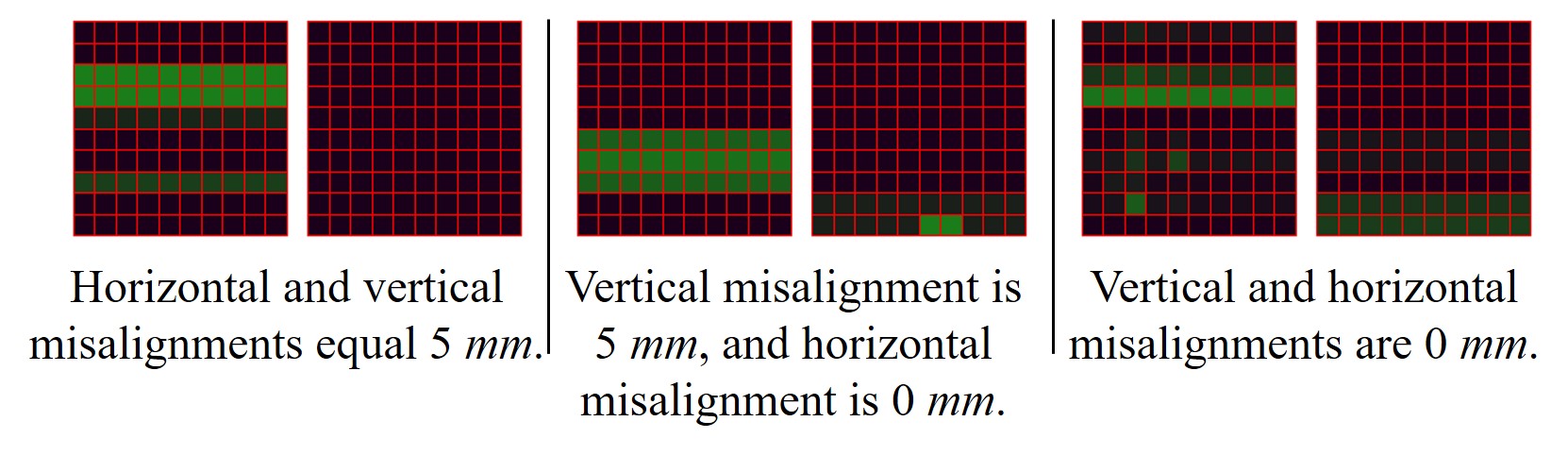}}
 \caption{Tactile patterns for two collected datasets. Green cells illustrate pressure points on the sensors. The color saturation of a cell represents the force applied to it: black cells correspond to 0 $N$ and bright green cells correspond to 9 $N$.}\label{PAT}
\end{center}
\vspace{-0.5em}
\end{figure}

The first dataset represents data on the angular misalignment between the electrodes. It includes 600 data pairs from tactile sensor arrays. The angle at which the main robot was tilted ranged from 0 to 5$^{\circ}$. For every angle span of 1$^{\circ}$, the end effector docked with the main robot, and values from 2 tactile sensors were recorded 100 times. The end effector starting position varied from -20 to 20 $mm$ along the $X$ and $Y$ axes with a step of 4 $mm$. The input data included 10x10 array with 2 channels, because there were two data pairs per sensor measurement. Fig. \ref{PAT}(a) presents a set of obtained patterns for different misalignment angles. 

The second dataset contains data on the horizontal and vertical misalignment of the end effector and includes 500 data pairs from tactile sensor arrays. The patterns of translational misalignments were collected for the case of zero angular misalignment. The actuator was installed opposite to the counterpart with electrodes so that the coordinates of the electrodes in the actuator coordinate system were equal to (0; 0; 80) $mm$. After that, the robot performed 20 attempts to dock with the electrodes. Similarly, the actuator's target point was changing with a 5 $mm$ increment along the $X$ and $Y$ axes. As a result, the coordinates of the end effector changed from (-10; -10; 80) $mm$ to (10; 10; 80) $mm$. The example of collected tactile patterns is shown in Fig. \ref{PAT}(b).

% \begin{comment}
% Before implementing the CNN in our tactile perception system, we compared different shallow and deep learning (regular neural network (NN) and CNN) approaches for the classification problem. The shallow learning approaches included the following classification models: 
% \begin{enumerate}
% \item Logistic Regression (LR) Classifier with cross-validation estimator,
% \item Support  Vector  Machine (SVM) with  stochastic  gradient descent learning, 
% \item K-nearest neighbors (KNN) Classifier,
% \item Decision Tree (DT) Classifier,
% \item Gradient Boosted Decision Tree (GBDT) Classifier, 
% \item  Random Forest (RF) Classifier,
% \item  XGBoost Classifier.
% \end{enumerate}

%  \begin{figure}[ht]
%   \centering
%      \includegraphics[width=1\linewidth]{figures/Experiments/Angular.png}
%   \caption{Accuracy comparison of all explored classification models for the angle prediction.}
%      \label{Accuracy}
% \end{figure}
% \end{comment}

To avoid resubstitution validation or evaluation we split the data into training (67\%) and validation sets (33\%) \cite{raschka2018model}. The collected datasets were balanced, therefore, we chose accuracy as a quality metric for the evaluation of our model. The performance of applied CNN for the misalignment prediction is shown in Table \ref{tabl:T2}.

\begin{table}[!h]
    \centering
    \caption{Accuracy of Applied CNN for the Misalignment Prediction}
    \label{tabl:T2}
    \begin{tabular}{|c|c|c|c|}
    \hline
    \multirow{2}{*}{CNN} & \multicolumn{3}{c|}{\shortstack{\\Misalignment case}}\\
    \cline{2-4} 
     & Angular & Vertical & \shortstack{\\Horizontal} \\
      \hline
    Accuracy &  0.955 &  0.982 & \shortstack{\\0.869}\\
    \hline
    \end{tabular}
\end{table}

\section{Experiments}
All object detection models were trained on the Intel NUC Mini PC (parameters are listed in the Table \ref{tabl:T3}).

\begin{table}[ht]
    \centering
    \caption{Parameters of the Computer}
    \label{tabl:T3}
    \scalebox{0.88}{%
    \begin{tabular}{|c|c|}
    \hline
     Computing Unit &  \shortstack{\\Intel NUC NUC7i5BNK} \\
    \hline
    \multirow{2}{*}{Processor} &  \shortstack{\\Intel Core i5-7260U} \\ 
     & (2.7 GHz – 3.4 GHz, Quad Core, 4M Cache, 15W TDP) \\
    \hline
     Memory & \shortstack{\\Two DDR4 SO-DIMM sockets (32 GB, 2133 MHz)}   \\
    \hline
     Operating system &  \shortstack{\\Ubuntu 20.04.2 LTS} \\
    \hline
    \end{tabular}}
\end{table}

\subsection{Comparison of computer vision techniques for the electrode position detection}

We have explored 3 different computer vision (CV) techniques for the electrode position detection problem: 
\begin{enumerate}
   
	\item Contour detection: we drop the pixels of an image, using HSV filter. After that, we applied Canny edge detector \cite{CANNY} to find the edges and to build different contours. If a contour with the area is more than the area's threshold, it is considered as the contour of electrodes. The method is straightforward but not robust.
    \item Histogram of oriented gradients (HoG) with Support Vector Machine (SVM) \cite{HoG}: in this method, we apply a feature descriptor to train a machine learning model to detect the electrodes. HoG is the number of occurrences of gradient orientation in localized portions of an image.
     \item Deep learning approach: CNN YOLOv3 retrained on our dataset.
    
\end{enumerate}

\begin{figure}[!h]
 \begin{center}
 \subfigure[Comparison of applied CV techniques.]{
 \includegraphics[width=0.95\linewidth]{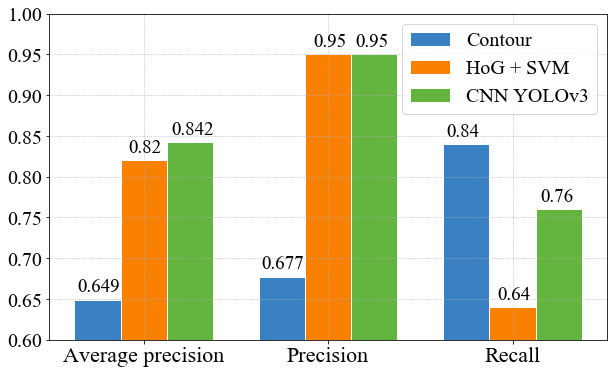}}
 \subfigure[The execution time of the explored CV techniques.]{
 \includegraphics[width=0.95\linewidth]{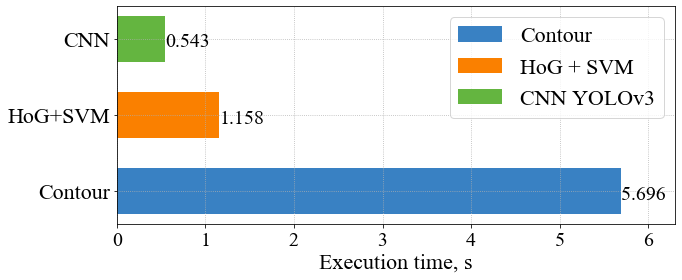}}
 \caption{The results of the algorithm reliability experiment.}\label{CV_performance_time}
\end{center}
\vspace{-1em}
\end{figure}

The test dataset included 50 samples, where the distance between the robots varied from 5 to 40 $cm$. We compared the performance of applied detection methods based on three metrics: Average Precision (AP), Precision, and Recall. AP is the area under the precision-recall curve. It is the main metric, which shows the accuracy of the object detection technique. In addition to the object detection metrics, we considered the execution time as the fourth metric. The execution time is the important parameter of any real-time system. In this work, we do not have the goal to achieve high speed of the docking process, but the speed of detection is still crucial for the future improvements of the system. The comparison results of explored CV methods are presented in Fig. \ref{CV_performance_time}.

CNN has a better performance than HoG and requires less time on object detection. CNN generally has a higher execution time than techniques without complex architectures. However, in our case, the execution time of CNN is two times lower than HOG. It makes CNN the best CV technique trained on our dataset to use during the docking process. 

\subsection{Reliability of the electrode search algorithm}

\begin{figure}[h]
\centering
\includegraphics[width=0.92\linewidth]{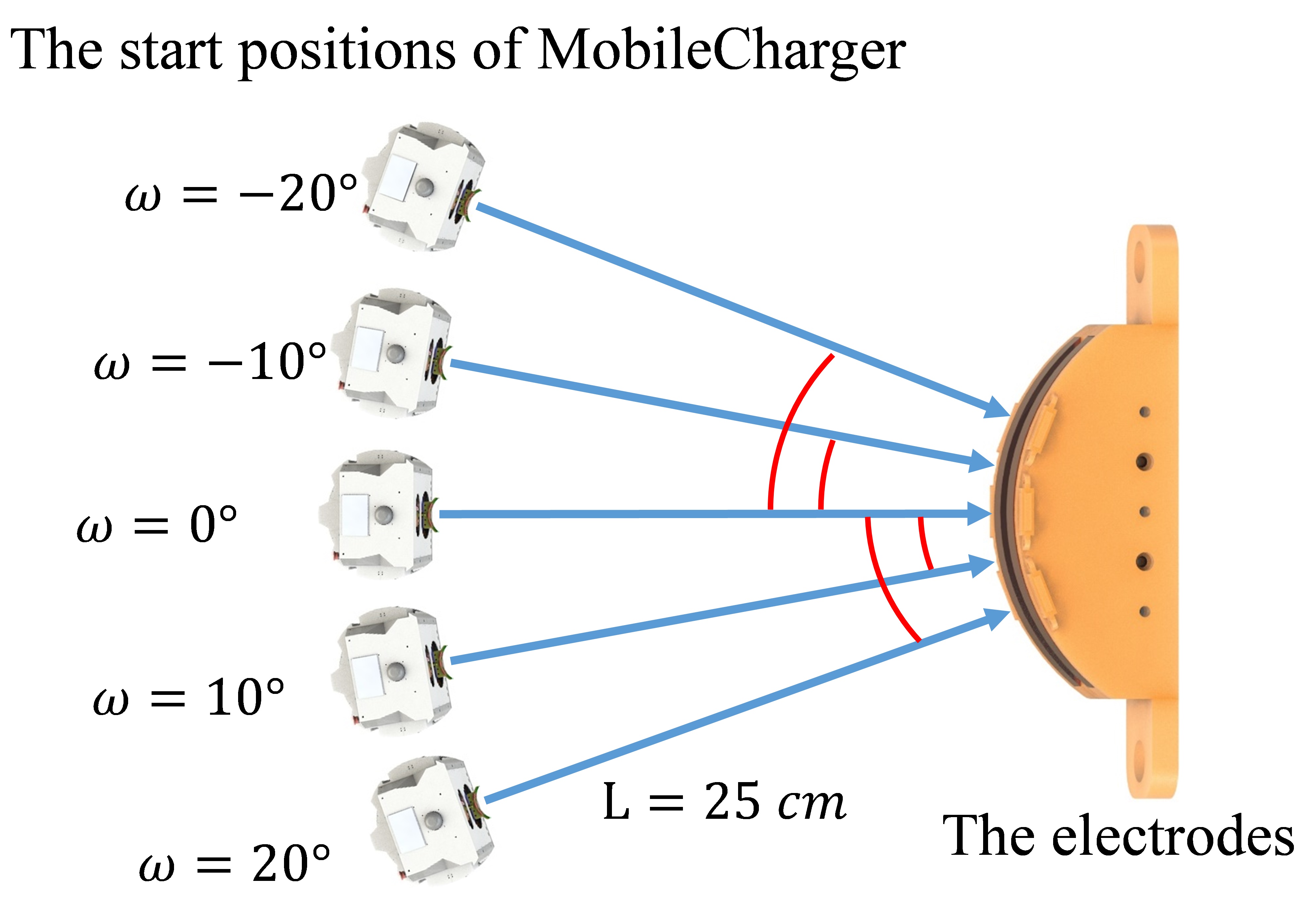}
\caption{The experimental setup. $\omega$ is the angle of robot rotation (yaw angle) with respect to the electrodes. $L$ is the distance between the initial position of MobileCharger and counterpart with electrodes.}
\label{experimentalSetup}
\end{figure}

To test the robustness of the electrode search algorithm (Algorithm \ref{algorithm}), we installed the counterpart with electrodes on the stand and launched the charging robot from different positions. The starting positions of MobileCharger are shown in Fig. \ref{experimentalSetup}. 

\begin{figure}[!h]
 \begin{center}
 \subfigure[\label{CV_performance}The percentage of successful trials for different robot positions.]{
 \includegraphics[width=0.95\linewidth]{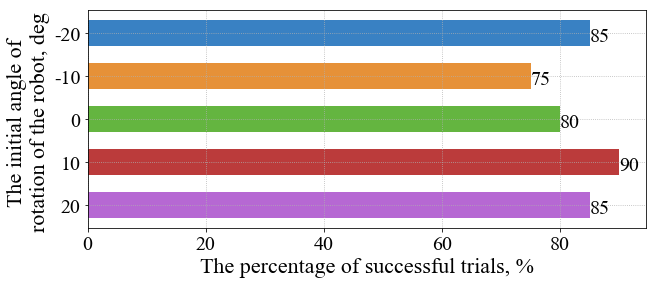}}
 \subfigure[\label{CV_time}The execution time of the algorithm. Mean values are marked with crosses.]{
 \includegraphics[width=0.95\linewidth]{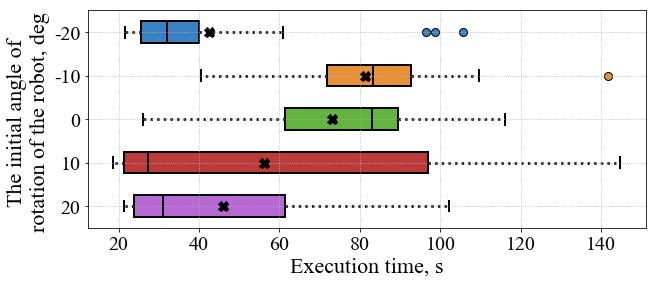}}
 \caption{The results of the algorithm reliability experiment.}\label{barplotTime}
\end{center}
\vspace{-1em}
\end{figure}

There were 5 initial positions for MobileCharger differing in the angle of rotation relative to the target electrodes $\omega$: -20$^{\circ}$, -10$^{\circ}$, 0$^{\circ}$, 10$^{\circ}$, 20$^{\circ}$. The height of the stand was 16 $cm$. The distance between the starting position of MobileCharger and electrodes $L$ was 25 $cm$, as it is the maximum distance for a given height of the stand at which the camera can detect the electrodes. In total, 100 trials were carried out (20 trials for each initial position). To evaluate the algorithm performance, we calculated its execution time and percentage of successful trials. The results of the experiment are presented in Fig. \ref{barplotTime}. 

The average percentage of successful trials of the algorithm is 83$\%$. One-way ANOVA results showed that there is no statistically significant difference among the starting positions $(F=0.44,\ p=0.78 > 0.05)$. Thus, the success of the algorithm operation is independent of the initial orientation of the robot. In general, the execution time of the algorithm varied from 19 to 145 $s$, and the average execution time of the algorithm for all trials was 60 $s$, SD=33 $s$. Using one-way ANOVA with a chosen significance level of $p<0.05$ we found a statistically significant difference in the speed of the algorithm execution $(F(4,95)=6.33,\ p=1.47 \cdot 10^{-4}<0.05)$. According to ANOVA results, the algorithm execution time for the initial position of the robot rotated by -20$^{\circ}$ was statistically significantly lower than for the  positions with $\omega=0^{\circ}$ $(F(1,38)=14.11,\ p=5.8 \cdot 10^{-4}<0.05)$ and $\omega=-10^{\circ}$ $(F(1,38)=22.69,\ p=2.8 \cdot 10^{-5}<0.05)$. Similarly, the execution time for the robot positioned with $\omega=20^{\circ}$ was statistically significantly lower than for the initial positions with $\omega=0^{\circ}$ $(F(1,38)=10.4,\ p=2.6 \cdot 10^{-3}<0.05)$ and $\omega=-10^{\circ}$ $(F(1,38)=17.65,\ p=1.6 \cdot 10^{-4}<0.05)$. Therefore, we can conclude that the larger angle of initial robot rotation leads to the faster execution of the algorithm. This can be explained by the fact that when the full image of the electrodes is presented in the RGB image, the algorithm requires more time to correct the robot path.

\section{Conclusions}

In this work, we presented MobileCharger, the mobile charging robot equipped with vDelta mechanism and high-fidelity tactile sensors. We invented vDelta to achieve a reliable and compact structure for the charging actuator. MobileCharger uses the RGB-D camera-based CV system to detect the electrodes on the target robot under various lighting conditions. CNN allowed to achieve the average precision of 84.2\% in electrode detection. We tested the robustness of the electrode search algorithm performed with CNN, which showed a good result of 83 successes out of 100. The mean execution time of the algorithm was 60 $s$. The proposed CNN-driven tactile perception system can accurately estimate the angular, vertical, and horizontal values of end effector misalignment from the pressure data. The estimated angle allows the system to decide whether the current position of the electrodes is safe for charging or there is a risk of a short circuit.
The presented mobile charger can be applied in indoor and outdoor environments. MobileCharger can increase the operation time of mobile robots and increase the level of autonomy of the entire robotic systems.

In future work, we plan to develop collision avoidance, localization, and mapping algorithms for the presented mobile robot. Furthermore, we are going to improve the end effector mechanism by adding 1-DoF for electrode rotation around the roll axis. This would allow charging the main robot tilted at the critical angle. The future system SwarmCharge will include swarm of charging and mobile robots with optimal charging scheduling and path planning. Additionally, we will make it possible to charge heterogeneous robots, e.g., quadruped robots, humanoid robots, landing UAV, and etc.

\section*{Acknowledgements}

The authors would like to thank Professor Hiroyuki Kajimoto, University of Electro-Communications, Japan for providing tactile sensor arrays, and researcher Jonathan Tirado, Skolkovo Institute of Science and Technology, for helping with setting up the tactile sensors.

\bibliographystyle{ieeetr} 
\bibliography{references}

\end{document}